%% file: main.tex
\documentclass[preprint,5p,times,twocolumn]{elsarticle}

\usepackage{xcolor}
\usepackage{graphicx}
\usepackage{amsmath}
\usepackage{amssymb}
\usepackage{booktabs}
\usepackage{float}
\usepackage{subcaption}
\usepackage{multirow}

\usepackage{pifont}%
\newcommand{\cmark}{\ding{51}}%
\newcommand{\xmark}{\ding{55}}%

\usepackage[capitalize]{cleveref}
\crefname{section}{Sec.}{Secs.}
\Crefname{section}{Section}{Sections}
\Crefname{table}{Table}{Tables}
\crefname{table}{Tab.}{Tabs.}

\journal{Signal Processing: Image Communication}

\begin{document}
\begin{frontmatter}

\title{ProtoSeg: A Prototype-Based Point Cloud Instance Segmentation Method}

\author{Remco Royen\corref{cor1}}
\ead{rroyen@vub.be}
\author{Leon Denis}
\author{Adrian Munteanu}
\cortext[cor1]{Corresponding author}

\address{Department of Electronics and Informatics, Vrije Universiteit Brussel, 1050 Brussels, Belgium\\
imec, 3001 Leuven, Belgium}

\begin{abstract}
3D instance segmentation is crucial for obtaining an understanding of a point cloud scene. This paper presents a novel neural network architecture for performing instance segmentation on 3D point clouds. We propose to jointly learn coefficients and prototypes in parallel which can be combined to obtain the instance predictions. The coefficients are computed using an overcomplete set of sampled points with a novel multi-scale module, dubbed dilated point inception. As the set of obtained instance mask predictions is overcomplete, we employ a non-maximum suppression algorithm to retrieve the final predictions. This approach allows to omit the time-expensive clustering step and leads to a more stable inference time. The proposed method is not only 28\% faster than the state-of-the-art, it also exhibits the lowest standard deviation. Our experiments have shown that the standard deviation of the inference time is only 1.0\% of the total time while it ranges between 10.8 and 53.1\% for the state-of-the-art methods. Lastly, our method outperforms the state-of-the-art both on S3DIS-blocks (4.9\% in mRec on Fold-5) and PartNet (2.0\% on average in mAP).
\end{abstract}


\end{frontmatter}



\input{texts/1_introduction.tex}
\input{texts/2_relatedworks.tex}
\input{texts/3_proposedmethod.tex}
\input{texts/4_experiments.tex}

\input{texts/5_conclusion.tex}

\section*{Acknowledgment}
The first author is a FWO-SB PhD fellow funded by Research Foundation - Flanders (FWO), project number 1S89420N.




 \bibliographystyle{elsarticle-num-names} 
 \bibliography{references}





\end{document}

%% file: texts/1_introduction.tex
\section{Introduction}
\label{sec:intro}

The field of 3D technology is attracting a large amount of academic and industrial interest, driven by the availability of economical 3D sensors and the advent of deep learning. 3D scene understanding is of critical importance for a large amount of application domains, such as virtual reality \cite{kourbane2022hybrid}, autonomous driving \cite{butt2022carl, denis2023gpu}, drone exploration \cite{koyun2022focus} and robotics \cite{chang2020vision}. To this end, a fundamental, but challenging sub-domain is 3D instance segmentation. While semantic segmentation attempts to obtain solely a semantic label for each point~\cite{zhao2021point, royen2023resscal3d, royen2024resscal3d++}, instance segmentation requires a higher level of scene understanding to segment objects and is able to differentiate objects of the same semantic class.

2D instance segmentation has been exhaustively studied in the literature~\citep{he2017mask, bai2017deep, kong2018recurrent, neven2019instance, bolya2019yolact, huang2019mask, wang2020solo, tian2020conditional, wang2020solov2, fang2021instances, ke2022mask, zhu2022sharpcontour, zhao2023weight}. However, point cloud instance segmentation poses additional difficulties due to the unstructured nature of point clouds, large variations in instance sizes and input coordinate scales, and possibly a high number of input points.

State-of-the-art methods~\citep{jiang2020pointgroup, he2021dyco3d, chen2021hierarchical, vu2022softgroup} adopt bottom-up approaches where firstly discriminative features, semantic labels and center offset vectors are computed. In a subsequent step, a clustering algorithm is employed to retrieve instance candidates of which the best are retained. While these methods have shown good performances, their inference is relatively slow due to a time-expensive clustering step. Furthermore, the required inference time shows a high variability as the time of the clustering and candidate selection network strongly depend on the scene, impeding online applications.

\begin{figure}[t]
    \centering
    \includegraphics[width=0.99\linewidth]{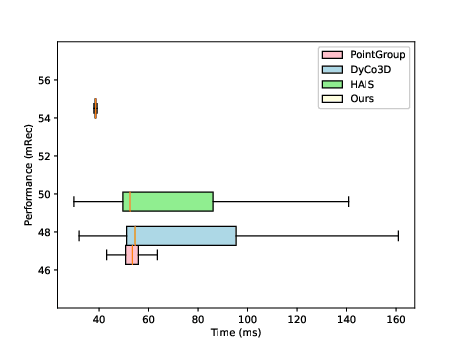}
    \caption{Speed-performance comparison on S3DIS-blocks Area-5. The proposed method outperforms the state-of-the-art in terms of accuracy, speed and variance in inference time.} 
    \label{fig:boxplot_speed_performance}
\end{figure} 

In order to address these issues, this paper proposes a novel, clustering-free approach that learns prototypes and coefficients in a joint manner. A linear combination of both allows to retrieve a set of instance mask candidates of which the best are retained using a simple and fast non maximum suppression (NMS) algorithm. While our method can technically be classified as a proposal-based method, in contrast to traditional proposal-based methods, we do not require precise proposal predictions as we simply sample a subset of points using farthest point sampling (FPS). By doing so, we avoid error accumulation and time-expensive proposal prediction. A novel Dilated Point Inception (DPI) module allows to retrieve multi-scale coefficients for each proposal. Important to note is that the set of proposals is overcomplete and has a fixed cardinality. A consequence of this approach is that the inference time is less dependent of the underlying scene and thus more predictable. Our experiments have shown that the standard deviation of the inference time on the test set is only 1.0\% of the total time while it ranges between 10.8 and 53.1\% for the state-of-the-art methods. Also, our method has shown to be the fastest overall with a reduction of 28\% in inference time compared to the current state-of-the-art. Additionally, our method outperforms the state-of-the-art both on S3DIS-blocks (4.9\% in mRec on Fold-5) and PartNet (2.0\% on average in mAP). A performance and speed comparison with the state-of-the-art is visualized with a boxplot in \cref{fig:boxplot_speed_performance}.


Summarised, our main contributions are as follows:
\begin{itemize}
    \item We propose a novel end-to-end prototype-based network architecture for 3D instance segmentation. Coefficients and prototypes are learnt jointly in parallel and are combined to obtain instance predictions. A reciprocal loss was specifically developed.
    \item A novel module, called Dilated Point Inception (DPI), allowing multi-scale coefficient retrieval and boosting performance. 
    \item The proposed work significantly outperforms the state-of-the-art for S3DIS-blocks and PartNet. Additionally, our overcomplete, clustering-free approach does not only achieve the fastest available inference but also offers the lowest variation in inference time.
    
\end{itemize}

\begin{figure*}[t]
    \centering
    \includegraphics[width=0.98\linewidth]{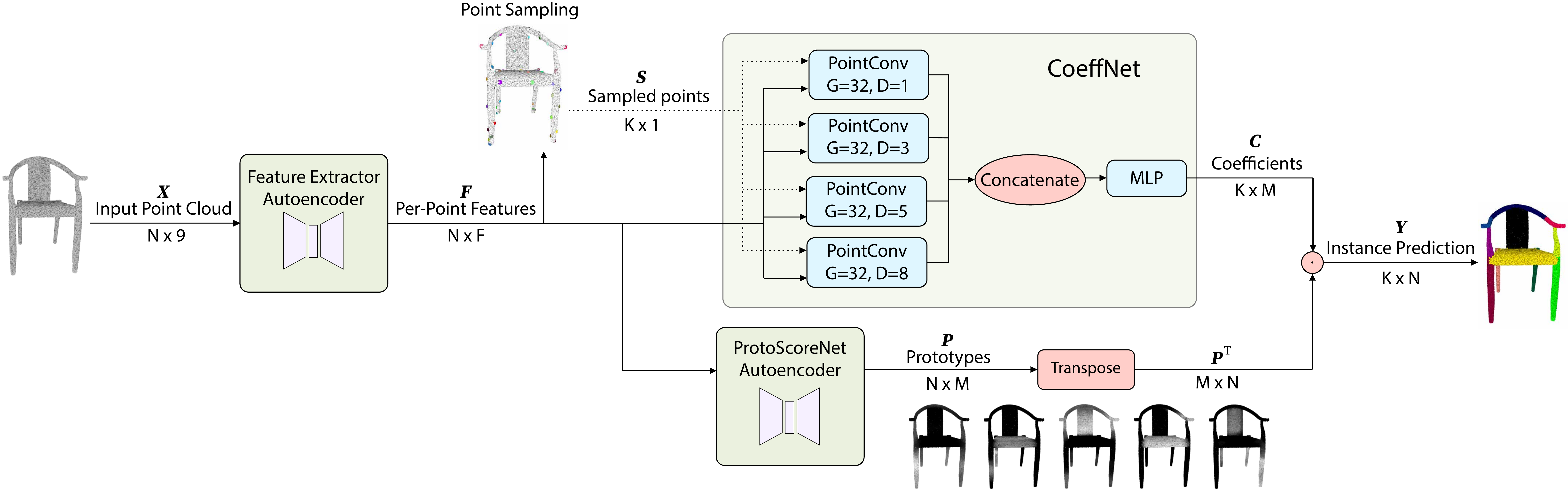}
    \caption{The ProtoSeg network architecture consists of four main parts: (1) A feature extractor which retrieves per-point features. (2) A point sampler, obtaining points with large diversity in feature space. (3) ProtoScoreNet, which retrieves a set of prototypes. In parallel, (4) Coeffnet computes for each sampled point a set of coefficients associated with the prototypes. To allow multi-scale coefficients retrieval, a Dilated Point Inception (DPI) module is employed. Instance predictions are obtained by linearly combining the coefficients and prototypes.}
    \label{fig:arch}
\end{figure*}

%% file: texts/2_relatedworks.tex
\section{Related work}

\noindent\textbf{Deep Learning on Point Clouds.}  Pioneering work in this domain was done by PointNet \citep{qi2017pointnet} as it was the first deep learning network that was able to operate on point clouds directly. To cope with the unstructured nature of point clouds, it employed permutation-invariant operators. In PointNet++ \citep{qi2017pointnet++}, these ideas were applied on local groups instead of on the whole point cloud. DGCNN \citep{wang2019dynamic} and PointWeb \citep{zhao2019pointweb} firstly construct a graph before employing graph network techniques. Other techniques such as PointConv \citep{wu2019pointconv} and KPConv \citep{thomas2019kpconv}, learn convolutional kernels which are dependant on the input coordinates. Recently, PointTransformer \citep{zhao2021point} applied the concept of self-attention layers for 3D point cloud processing leading to impressive results.

\noindent\textbf{Instance Segmentation.} With the advent of more powerful deep learning techniques, research attention has shifted to more challenging problems such as instance segmentation, both in 2D as 3D. The available literature can be divided into three categories: proposal-based, clustering-based and direct-segmentation methods. They are discussed next.

\textit{Proposal-based methods} split instance segmentation in two sequential problems namely, proposal and instance prediction. These proposals can come in different forms such as bounding boxes or center points for example. A popular approach in 2D is to obtain the former with an object detector, which is subsequently segmented to obtain the instance masks~\citep{he2017mask, huang2019mask, tian2020conditional, fang2021instances}. This top-down approach reduces instance segmentation to a combination of two well-investigated research domains. Yolact \citep{bolya2019yolact} follows a similar approach as it predicts coefficients for each detected bounding box and combines them with generated prototypes. The detect-then-segment paradigm has also been explored in 3D. 3D-BoNet \citep{yang2019learning} regresses a set of 3D bounding boxes which are combined with a parallel instance prediction branch. GSPN \citep{yi2019gspn} and GICN \citep{liu2020learning} determine regions of interest (RoI) by exploiting center prediction and center heatmaps respectively. In contrast to the previous methods, 3D-MPA \citep{engelmann20203d} does not employ bounding boxes as proposals. It takes an object-centric approach where each point votes for its object center, leading to different proposal groups. The obtained proposals are then further clustered to obtain the predicted instances. In NeuralBF \citep{sun2023neuralbf} an iterative
bilateral filtering is performed with learned kernels. To do so, it regresses the bounding hull and uses a semantic segmentation to construct an affinity function.

Traditionally, in proposal-based methods the predicted proposals are of critical importance. They inherently already provide a high amount of instance information. Thus, a sub-optimal set of proposals has a damaging effect on the output. The stacking of two tasks inevitably leads to accumulating errors, even when trained end-to-end.

\textit{Clustering-based methods} adopt a bottom-up approach. They mostly perform discriminative feature learning followed by a time expensive clustering algorithm. Several works have followed this point of view for 2D instance segmentation \citep{bai2017deep, kong2018recurrent, neven2019instance}. For 3D, the clustering-based approach is commonly used as the spatial separation between different instances allows performant clustering~\citep{wang2018sgpn, wang2019associatively, zhao2020jsnet, zhang2021point, chen2022jspnet, elich20193d, pham2019jsis3d, he2020instance}. SGPN \citep{wang2018sgpn} computes a similarity matrix which is used to merge similar points to instances. ASIS \citep{wang2019associatively}, JSNet \citep{zhao2020jsnet} and JSPNet \citep{chen2022jspnet} propose to perform semantic and instance segmentation jointly by introducing specifically designed modules connecting both branches. The resulting feature vector is used as input for a grouping algorithm. \citet{zhang2021point} suggest to represent each point as a tri-variate normal distribution and to use a novel loss function in the clustering. With MP-Net \citep{he2020learning}, a memory-augmented network was introduced that learns and memorizes representative feature prototypes. During training, a fixed dictionary of these prototypes can be used to augment the retrieved features and improve performance on non-dominant classes. While the above methods employ a block-based approach, \citep{jiang2020pointgroup, han2020occuseg, he2021dyco3d, chen2021hierarchical} process the full point-cloud directly. PointGroup \citep{jiang2020pointgroup} learns offsets which are used to increase intra-instance distances and aid the clustering. With a non-maximum suppression (NMS) and a dedicated network that attributes scores to each obtained candidate, redundant clusters are removed. In DyCo3D \citep{he2021dyco3d} dynamical convolutions are employed. A hierarchical aggregation is proposed in HAIS \citep{chen2021hierarchical} such that instance proposals are progressively generated. SoftGroup \citep{vu2022softgroup} further improves upon this work by introducing soft semantic labelling prior to grouping.

An important drawback of current clustering-based methods is the time-varying and time-expensive nature of their clustering algorithms.
These methods have recently shown to outperform the proposal-based methods, albeit part of the performance increase should be contributed to the full-scene processing done in the latest papers, as will be shown in \cref{subsec:exp_s3dis}.

\textit{Direct-segmentation.} For 2D images, SOLO \citep{wang2020solo} and its extension SOLOv2 \citep{wang2020solov2}, have presented a new approach, namely direct-segmentation without proposals nor clustering algorithms. Instead, the image is partitioned in fixed regions for which instance masks are predicted. By doing so, they obtain an overcomplete set of predictions without having to predict accurate proposal regions first. A NMS algorithm is used to remove the duplicates. To the best of our knowledge, no counterpart exists for 3D instance segmentation.

The  method proposed in this work can be regarded as a combination of the proposal-based and direct-segmentation approaches and is a significant extension of our prior work in \citep{royen2024joint}. While we sample proposal points, and thus technically we are a proposal-based method, the retained points do not represent a prediction of any sort and do not contain instance information. Points are sampled using FPS in feature space such that their surrounding local structures are diverse. After combination with the generated prototypes, the best instance predictions are retained. By doing so, an accumulation of errors and the usage of a clustering algorithm are avoided.

%% file: texts/3_proposedmethod.tex
\section{Proposed Method}
 
\subsection{Overview of the proposed method}

The architecture of the proposed method is illustrated in \cref{fig:arch} and consists out of 4 main parts. Firstly the input data $\boldsymbol{X} \in \mathbb{R}^{N \times I}$, with $N$ and $I$ the number of input points and channels, is processed by a feature extractor to obtain per-point features $\boldsymbol{F} \in \mathbb{R}^{N \times F}$, with $F$ the feature dimension. These features are the input of 2 parallel branches to compute the coefficients and prototypes. The former is obtained by firstly sampling $K$ points $\boldsymbol{S} \in \mathbb{R}^{K}$ with a dedicated algorithm (\cref{subsec:keypoint_select}) which is subsequently employed by CoeffNet (\cref{subsec:coeffnet}) to compute coefficients $\boldsymbol{C} \in \mathbb{R}^{K \times M}$, with $M$ the number of prototypes. In the parallel branch, ProtoScoreNet (\cref{subsec:protoscorenet}) is used to retrieve the prototypes $\boldsymbol{P} \in \mathbb{R}^{N \times M}$. A matrix multiplication between $\boldsymbol{P}^{T}$ and $\boldsymbol{C}$ yields the overcomplete set of instance mask predictions $\boldsymbol{Y} \in \mathbb{R}^{K \times N}$. Mathematically, the combination of the coefficients and prototypes can be expressed as follows,
\begin{align}
\boldsymbol{y}_{k}^{q} = \sum_{i=1}^{M} c_{k, i}^{q} \boldsymbol{p}_{i}^{q},
\end{align}
with $\boldsymbol{y}_{k}^{q} = \boldsymbol{y}_{k}(\boldsymbol{X}_{q}) \in \mathbb{R}^{N}$ the $k^{th}$ prediction mask in $\boldsymbol{Y}$ for input sample $\boldsymbol{X}_{q}$; $c_{k,i}^{q} = c_{k,i}(\boldsymbol{X}_{q}) \in \mathbb{R}$ the $i^{th}$ coefficient for the k-th sampled point and $\boldsymbol{p}_{i}^{q} = \boldsymbol{p}_{i}(\boldsymbol{X}_{q}) \in \mathbb{R}^{N}$ the $i^{th}$ prototype.

During inference, $\boldsymbol{Y}$ is thresholded and the redundant, inaccurate predictions are filtered out by a non-maximum suppression (NMS) algorithm. The resulting matrix of instance masks is denoted by $\boldsymbol{Y'} \in \mathbb{R}^{K' \times N}$.

\subsection{Point Sampling}
\label{subsec:keypoint_select}

\begin{figure}[b]
  \centering
   \includegraphics[width=0.99\linewidth]{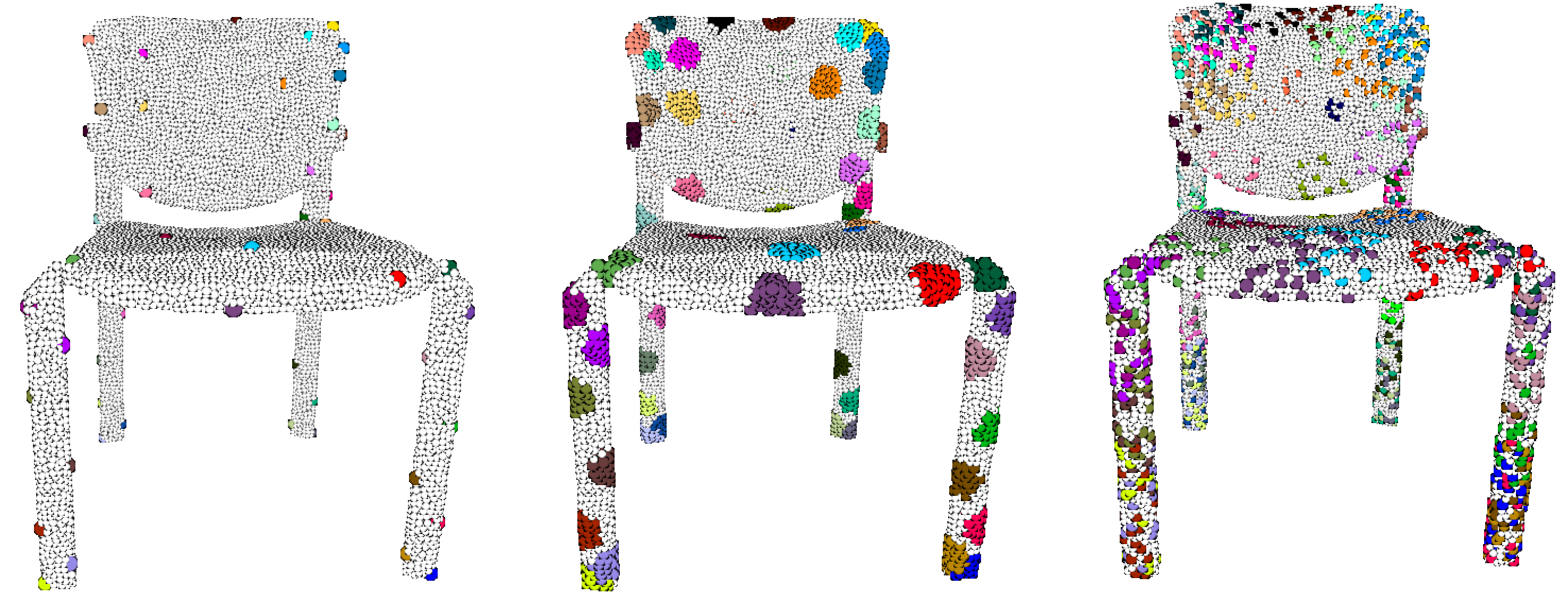}

   \caption{The different scales of DPI. From left to right: the sampled points and the receptive fields for the DPI branch with dilation factor 1 and 8, respectively.}
   \label{fig:multi_scale_receptive_field}
\end{figure}

In contrast to traditional proposal-based methods, we do not require accurate proposals such as 3D bounding boxes or center points, but we attempt to find a diverse set of sampled points $\boldsymbol{S}$. This diverse set of points allows CoeffNet (\cref{subsec:coeffnet}) to learn coefficients $\boldsymbol{C}$ from different local neighborhoods. Consequently, a large variation in combinations and instance proposals $\boldsymbol{Y}$ can be achieved. During inference, the best instance predictions $\boldsymbol{Y'}$ are selected from the overcomplete set $\boldsymbol{Y}$. This relaxed conditioning of the proposal points is an important difference with the traditional proposal-based points. By doing so, we set a first step towards a proposal-free, clustering-free 3D instance segmentation algorithm. We have adopted farthest point sampling (FPS) to sample $\boldsymbol{S}$ as it allows to sample points with diverse local surroundings, especially when performed on the features.

\begin{figure}[!b]
        \centering
        \includegraphics[width=\linewidth]{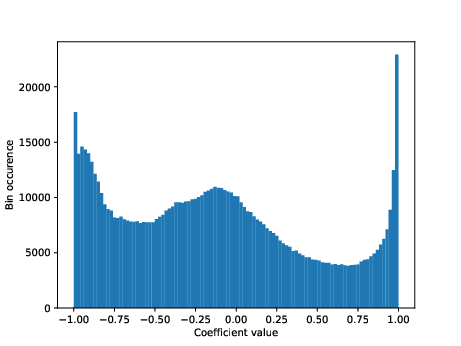}
     \caption{Distribution of the computed coefficients by ProtoSeg on the test set of PartNet level-1. Only the coefficients leading to an instance which is retained by NMS, are taken into account.}
    \label{fig:coef_histogram}
\end{figure}

\subsection{CoeffNet - Dilated Point Inception Module}
\label{subsec:coeffnet}

Given the set of sampled points $\boldsymbol{S}$ and the per-point features $\boldsymbol{F}$, CoeffNet computes a set of coefficients for each sampled point $\boldsymbol{S_{i}}$. To do so, it analyses the local structure around each $\boldsymbol{S_{i}}$. As $\boldsymbol{S}$ was selected in a way to favor the diversity of these neighborhoods, large variations in coordinates are found.
To enable the combination of multi-scale information around the sampled point, we introduce a novel Dilated Point Inception (DPI) module, depicted in the CoeffNet block in \cref{fig:arch}. In \citep{engelmann2020dilated}, Engelmann et al. presented dilated point convolutions. While it significantly increases the receptive field, it employs a fixed dilation factor and thus works on a single scale. With DPI, we propose an Inception-alike solution that combines different dilation factors. Convolution kernels are learnt at different scales as nonlinear functions of the local coordinates using four parallel branches with PointConv layers operating at different dilation factors. A visualization of the different scales can be found in \cref{fig:multi_scale_receptive_field}. After a concatenation, a MLP is employed to fuse these multi-scale features. A $tanh$ activation function is used to retrieve the coordinates as we have empirically noticed that the performance is positively influenced when coordinates are allowed to be negative. Stated differently, performance increases when prototypes are allowed to cancel regions of each other out. In \cref{fig:coef_histogram}, the distribution of the obtained coefficients on PartNet level-1 is visualised with a histogram. It can be noticed that the distribution exhibits three peaks, around approximately -1, 0 and 1. This stems from the employment of the $tanh$ activation function. It also shows that for each instance prediction a large amount of prototypes actively contribute. While some prototypes are used to highlight a specific zone with a coefficient close to 1, others are used to cancel regions with a coefficient close to -1.




\subsection{ProtoScoreNet}
\label{subsec:protoscorenet}
\begin{figure}[t]
  \centering
   \includegraphics[width=0.99\linewidth]{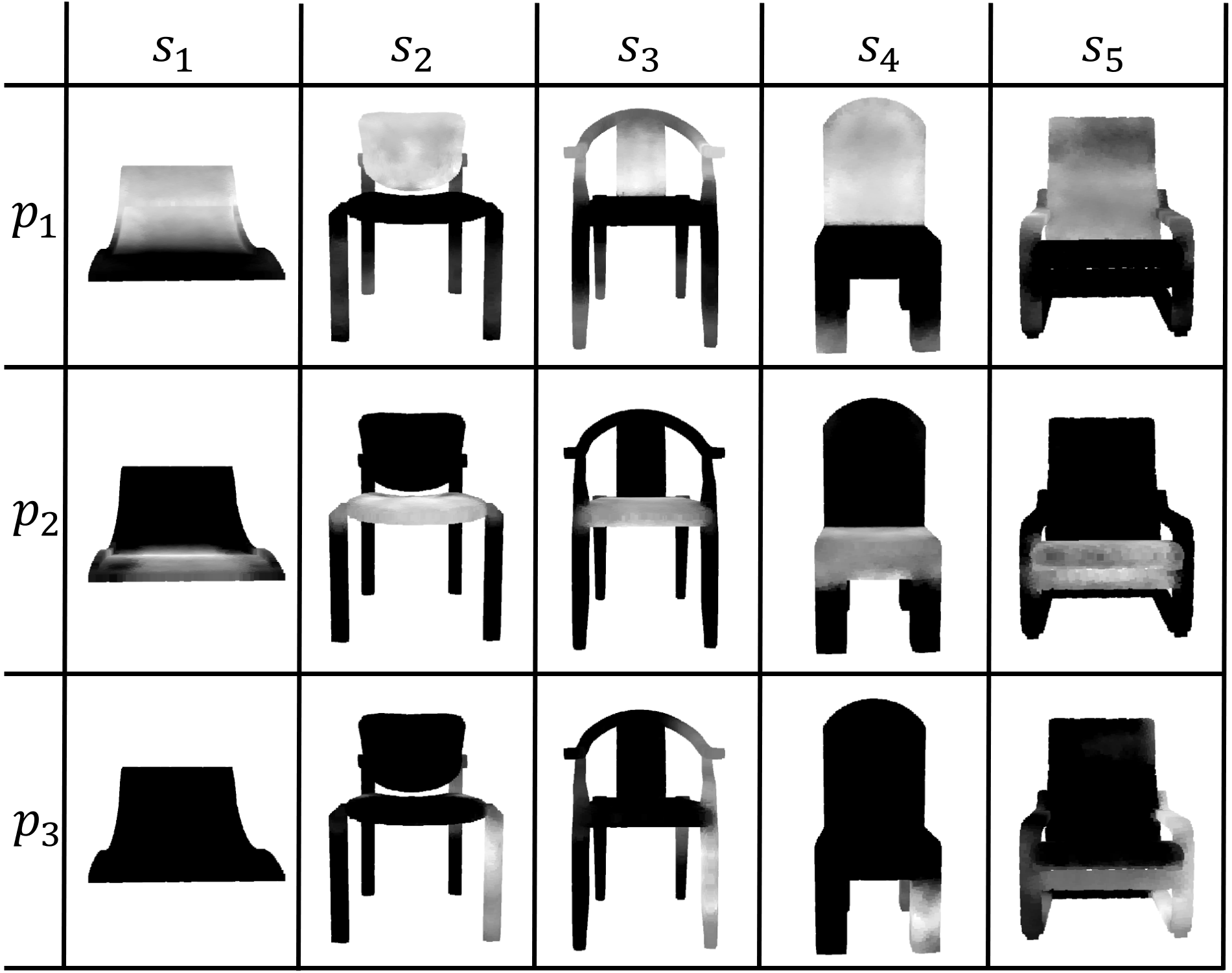}

   \caption{Prototype $p_i$ activation for different PartNet samples $s_i$.}
   \label{fig:proto_vis}
\end{figure}

Parallel to learning coefficients, prototypes $\boldsymbol{P}$ are generated. To ensure better convergence, CoeffNet and ProtoScoreNet are given features $\boldsymbol{F}$ from a shared feature extractor. The set of prototypes $\boldsymbol{P} \in \mathbb{R}^{N \times M}$ contains M prototypes $\boldsymbol{p} \in \mathbb{R}^{N}$, which are essentially vectors containing scores. As the point dimension is maintained throughout the prototype generation pipeline, these scores can be associated with the input point cloud and can thus be visualized spatially. Visual examples on PartNet are given in \cref{fig:proto_vis}. This shows that each prototype, that is, each column of matrix $\boldsymbol{P}\in \mathbb{R}^{N \times M}$, extracts specific information of the input. White and black points indicate a high and low score in a given prototype, respectively. If a given feature captured by a prototype is not present, all points will be given a zero score. As each prototype consists of N scores, the prototypes are defined over the whole point cloud, not per point. Note that the obtained prototypes clearly accentuate certain instance regions of the input point cloud and contain instance information. Linear combinations of such prototypes allows us to compute the instance predictions. 

\subsection{Loss-function}
\label{subsec:loss_function}

The proposed method is trained in an end-to-end manner. The output during training, $\boldsymbol{Y}$, is an overcomplete set of instance mask predictions. In contrast to semantic labels, instance numbers do not have a specific meaning. That is, permutations of the instance identifiers do not alter the results. Because of this, correspondences between the predicted instance and ground-truth (GT) masks must be derived prior to loss computation. A possible solution is to select the GT samples that are closest to each prediction. Mathematically, this can be expressed as
\begin{align}
  J_{PR \rightarrow GT}' = \sum_{i=1}^{I_{PR}} min_{j} L_{BCE}(\boldsymbol{Y_{i}}, \boldsymbol{GT_{j}}),
\label{eq:bad_loss_function}
\end{align}
with $j \in [1, I_{GT}]$ and where $I_{PR}$ and $I_{GT}$ are the number of instances in the prediction and ground-truth, respectively, $\boldsymbol{Y_{i}}$ is the $i^{th}$ predicted mask, $\boldsymbol{GT_{j}}$ the $j^{th}$ ground-truth mask and $L_{BCE}$ the binary cross-entropy loss. However, the usage of \cref{eq:bad_loss_function} leads to sub-optimal results. The main reason is that the selection of ground-truth labels is solely dependent of the network's own output. Thus, independent of the quality of the network's output, it receives a positive reinforcement as the losses are actively minimized. Hence, training becomes unstable. To solve this, a spatial constraint is added by using the sampled points. As each sampled point $S_{i} \in \boldsymbol{S}$ leads to a single instance mask prediction $\boldsymbol{Y_{i}}(S_{i}) \in \boldsymbol{Y}$ but at the same time also belongs to a ground-truth instance, this correspondence can be used to match predictions with the their ground-truth in an unambiguous manner. This is expressed formally as:
\begin{align}
  J_{PR \rightarrow GT} = \sum_{i=1}^{I_{PR}} L_{BCE}(\boldsymbol{Y_{i}}(S_{i}), \boldsymbol{GT}(S_{i})), 
\end{align}
with $\boldsymbol{GT}(S_{i})$ the ground-truth sample to which $S_{i}$ belongs. Experimental results have proven that this spatial constraint improves convergence. While $J_{PR \rightarrow GT}$ will ensure a GT instance close to each prediction and thus improve precision, it does not impose a predicted instance close to each GT. This inverse statement is expressed mathematically as follows:
\begin{align}
  J_{GT \rightarrow PR} = \sum_{i=1}^{I_{GT}} min_{j} L_{BCE}(\boldsymbol{GT_{i}}, \boldsymbol{Y_{j}}),
\end{align}
and will improve recall. The loss function we introduce and employ to train our network is a combination of those two terms, dubbed reciprocal loss, and is expressed mathematically as:
\begin{align}
  J_{RL} = J_{PR \rightarrow GT} + \lambda J_{GT \rightarrow PR},
\end{align}
with $\lambda$ a parameter to balance the two terms.

%% file: texts/4_experiments.tex
\section{Experiments}
\label{sec:experiments}

To validate the performance of ProtoSeg, we conduct quantitative and qualitative experiments on 2 widely-used reference datasets: S3DIS \citep{armeni20163d} and PartNet \citep{mo2019partnet}. On both datasets state-of-the-art performance is achieved.

\subsection{Dataset and Evaluation metrics}
S3DIS \citep{armeni20163d}, ScanNetV2 \citep{dai2017scannet} and PartNet \citep{mo2019partnet} are three popular datasets providing (part) instance annotations. It can be noticed that methods that employ a block-based (BB) approach generally present their results on S3DIS~\citep{wang2018sgpn, wang2019associatively, zhao2020jsnet, he2020instance, he2020learning, chen2022jspnet} and only seldom also on ScanNetV2~\citep{he2020instance, he2020learning}. Since we will employ a BB approach for the indoor room scenes, as discussed in \cref{subsubsec:block_based}, we have opted to present our results on S3DIS instead of ScanNetV2 to facilitate comparison with other block-based approaches.

The Stanford 3D Indoor Scene dataset (S3DIS)~\citep{armeni20163d} consists of 6 large-scale indoor areas with in total 271 rooms. Each point has been annotated with one of the 13 semantic categories. We follow the same evaluation strategy as the literature~\citep{zhao2020jsnet}, namely both Area-5 and 6-fold cross-validation testing. Mean precision (mPrec), mean recall (mRec), mean coverage (mCov) and mean weighted coverage (mWCov) as defined in \citep{zhao2020jsnet} are used as evaluation metrics.

\begin{table*}[t]
\begin{center}
\begin{tabular}{|c|c c c c | c c c c|} 
    \cline{2-9}
    \multicolumn{1}{c|}{} & \multicolumn{4}{c|}{5th-fold} & \multicolumn{4}{c|}{6-fold CV}\\
    \hline
    Method & mCov & mWCov & mRec & mPrec & mCov & mWCov & mRec & mPrec\\ 
    \hline
    JSNet~\citep{zhao2020jsnet}               & 48.7 & 51.5 & 46.9 & 62.1 & 54.1 & 58.0 & 53.9 & 66.9\\ 
    IAM~\citep{he2020instance}                & 49.9 & 53.2 & 48.5 & 61.3 & 54.5 & 58.0 & 51.8 & 67.2\\ 
    MPNet~\citep{he2020learning}               & 50.1 & 53.2 & 49.0 & 62.5 & \bf{55.8} & 59.7 & 53.7 & 68.4\\
    PointGroup*~\citep{jiang2020pointgroup}  & 43.1 & 47.8 & 46.8 & 55.1 & 48.4 & 54.4 & 55.3 & 59.4\\
    DyCo3D*~\citep{he2021dyco3d}             & 44.4 & 50.0 & 47.8 & 59.3 & 47.9 & 54.5 & 52.4 & 58.9\\
    HAIS*~\citep{chen2021hierarchical}        & 45.4 & 50.6 & 49.6 & 58.4 & 48.2 & 55.4 & 55.0 & 60.7\\
    SoftGroup*~\citep{vu2022softgroup}        & 40.6 & 45.6 & 43.9 & 42.6 & 45.0 & 51.2 & 48.4 & 52.3\\
    JSPNet~\citep{chen2022jspnet}            & 50.7 & 53.5 & 48.0 & 59.6 & 54.9 & 58.8 & 55.0 & 66.5\\
    \hline
    Ours                                    & \bf{51.0} & \bf{57.1} & \bf{54.5} & \bf{66.1} & 53.3 & \bf{60.8} & \textbf{57.9} & \textbf{70.8}\\ 
    \hline
\end{tabular}
\caption{Instance segmentation result on S3DIS-blocks for both Area-5 and 6-fold cross validation. The methods denoted with * were retrained on blocks.}
\label{tab:s3dis}
\end{center}
\end{table*}


\begin{table*}
\centering
\begin{tabular}{|c|c|c|c||c|} 
    \hline
    Method & Network (ms) & NMS (ms) & Grouping (ms) & Total (ms) \\ 
    \hline
    JSNet~\citep{zhao2020jsnet}              & 34.0 $\pm$ 11.3 & / & 96.2 $\pm$ 70.4 & 133.6 $\pm$ 71.0\\ 
    PointGroup~\citep{jiang2020pointgroup}   & 47.1 $\pm$ 6.6 & 0.4 $\pm$ 0 & 4.8 $\pm$ 1.5 & 53.2 $\pm$ 6.3\\ 
    DyCo3D~\citep{he2021dyco3d}              & 63.3 $\pm$ 28.7 & 0.5 $\pm$ 0.7 & 6.5 $\pm$ 7.3 & 72.7 $\pm$ 35.4\\ 
    HAIS~\citep{chen2021hierarchical}        & 57.4 $\pm$ 23.3 & / & 9.1 $\pm$ 10.0 & 69.3 $\pm$ 30.5\\
    SoftGroup~\citep{vu2022softgroup}        & 43.8 $\pm$ 3.5 & / & 5.6 $\pm$ 2.7 & 55.4 $\pm$ 6.0\\
    \hline
    Ours & 37.8 $\pm$ 0.4 & 0.4 $\pm$ 0 & / & \textbf{38.4 $\pm$ 0.4}\\  
    \hline
\end{tabular}
\caption{Instance segmentation inference time and standard deviation on S3DIS-blocks Area-5. For fair comparison, all runtimes were measured on the same CPU and NVIDIA GeForce RTX 3090.}
\label{tab:inference_time}
\end{table*}

PartNet \citep{mo2019partnet}, on the other hand, is a large-scale dataset of 3D objects that provides coarse- (level-1), middle- (level-2) and fine-grained (level-3) instance level 3D part information. It consist out of 26671 3D models covering 24 object categories. Training, validation and test splits are provided for each category. We follow experimental settings of \citep{he2020learning} and present results on the 'Chair' category, for the three granularity levels.
Part-category mean Average Precision (mAP) with IoU threshold of 0.5 is used to evaluate the performance, following the literature~\citep{mo2019partnet, he2020learning, zhang2021point}.

\subsection{Implementation details}
We process S3DIS in a block-based manner, following the experimental settings of \citep{qi2017pointnet, zhao2020jsnet}. Thus, we divide each room in 1m $\times$ 1m blocks with stride 0.5 such that an overlap is created. In each block 4096 points are randomly sampled during training. Each input point is represented by a 9-dimensional vector: XYZ, RGB and the normalized location of the point in the whole room.
For PartNet, only XYZ information is available. Since the samples of this dataset consist out of a much smaller amount of points, 10000, they are processed as a whole.

For the experiments, we have used a PointNet \citep{qi2017pointnet} feature extractor, PointConv without densitynet \citep{wu2019pointconv} in CoeffNet and PointNet++ \citep{qi2017pointnet++} for ProtoScoreNet. Semantic segmentation is obtained, in a block-based manner, by a PointTransformer \citep{zhao2021point} network. We use an Adam optimizer \citep{kingma2014adam} with learning rate 0.001 and train the network with batch size 16 for 65 epochs. With the exception of a batch size of 10, the same settings are used for PartNet. We have empirically selected the number of prototypes, features, sampled points and the sampling method as 128, 64, 64 and 'FPS on features', respectively, as it led to the best results.


During inference, we employ a threshold of 0.3 and use a regular NMS to retrieve the instance mask predictions for each block. The same BlockMerge algorithm as proposed in \citep{wang2018sgpn} is employed for S3DIS to ensure a fair comparison between block-based methods.
For PartNet, points without predictions are attached to the nearest instances of the same semantic class in a subsequent postprocessing step.

\subsection{S3DIS Instance Segmentation}
\label{subsec:exp_s3dis}

\subsubsection{Block-based instance segmentation}
\label{subsubsec:block_based}

We start by evaluating ProtoSeg on S3DIS. In the literature, two main approaches to process large-scale and dense datasets such as S3DIS and ScanNetV2 exist. In \citep{wang2018sgpn, zhao2020jsnet, he2020learning, he2020instance, denis2023improved}, a block-based (BB) approach is employed where each room is firstly divided in blocks which are processed separately. In a subsequent step, the resulting segmented blocks are merged with the BlockMerge algorithm~\citep{wang2018sgpn, denis2023improved}. In \citep{jiang2020pointgroup, he2021dyco3d, chen2021hierarchical, vu2022softgroup} the scene is processed as a whole without the division in blocks. While the latter has the advantage of having a global overview and thus leads to a higher segmentation performance, it requires a large GPU memory size which is also dependant of the input point cloud size. The block-based approach, on the other hand, is able to achieve a constant complexity with regard of the input size and also requires a much smaller amount of GPU memory, enabling embedded applications. 

\begin{figure}[t]
\centering
\includegraphics[width=0.8\linewidth]{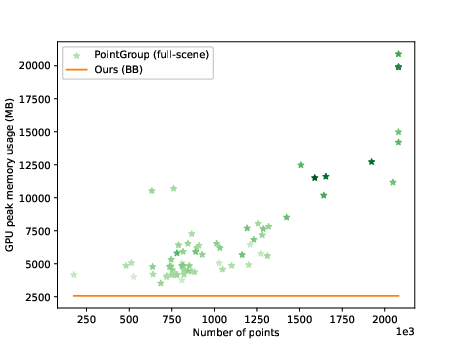}
\caption{Peak memory usage of the proposed method (block-based) and PointGroup \citep{jiang2020pointgroup} (full-scene) for different number of input points and number of instances (intensity of marker). The experiment is conducted on the scenes of S3DIS Area-5 \citep{armeni20163d}}
\label{fig:mem_reqs}
\end{figure}

To support this claim, we have evaluated the peak-memory usage of the GPU during inference using a full-scene approach (PointGroup \citep{jiang2020pointgroup}) and a BB approach (the proposed method) for the rooms in Area-5 of S3DIS. The results are presented in \cref{fig:mem_reqs} and show that a GPU peak memory of 2.6GB is enough to process point clouds of arbitrarily large size with our method while the required complexity of PointGroup rises with additional points to over 20GB. It should be noted that even on our performant NVIDIA GeForce RTX 3090 GPU, we were unable to process four of the largest scenes with PointGroup and had to downsample them to obtain results. BB-methods are also suitable for online applications where the scene has to be processed during capturing. As a consequence, block-based approaches, even though achieving a lower performance, remain relevant. Due to computational constraints, we were unable to apply ProtoSeg on the scene as a whole and we will present our results only with a BB-approach (S3DIS-blocks).


\subsubsection{Results}


\begin{figure*}
    \centering
    \begin{subfigure}[b]{0.3\textwidth}
        \centering
        \includegraphics[width=\textwidth]{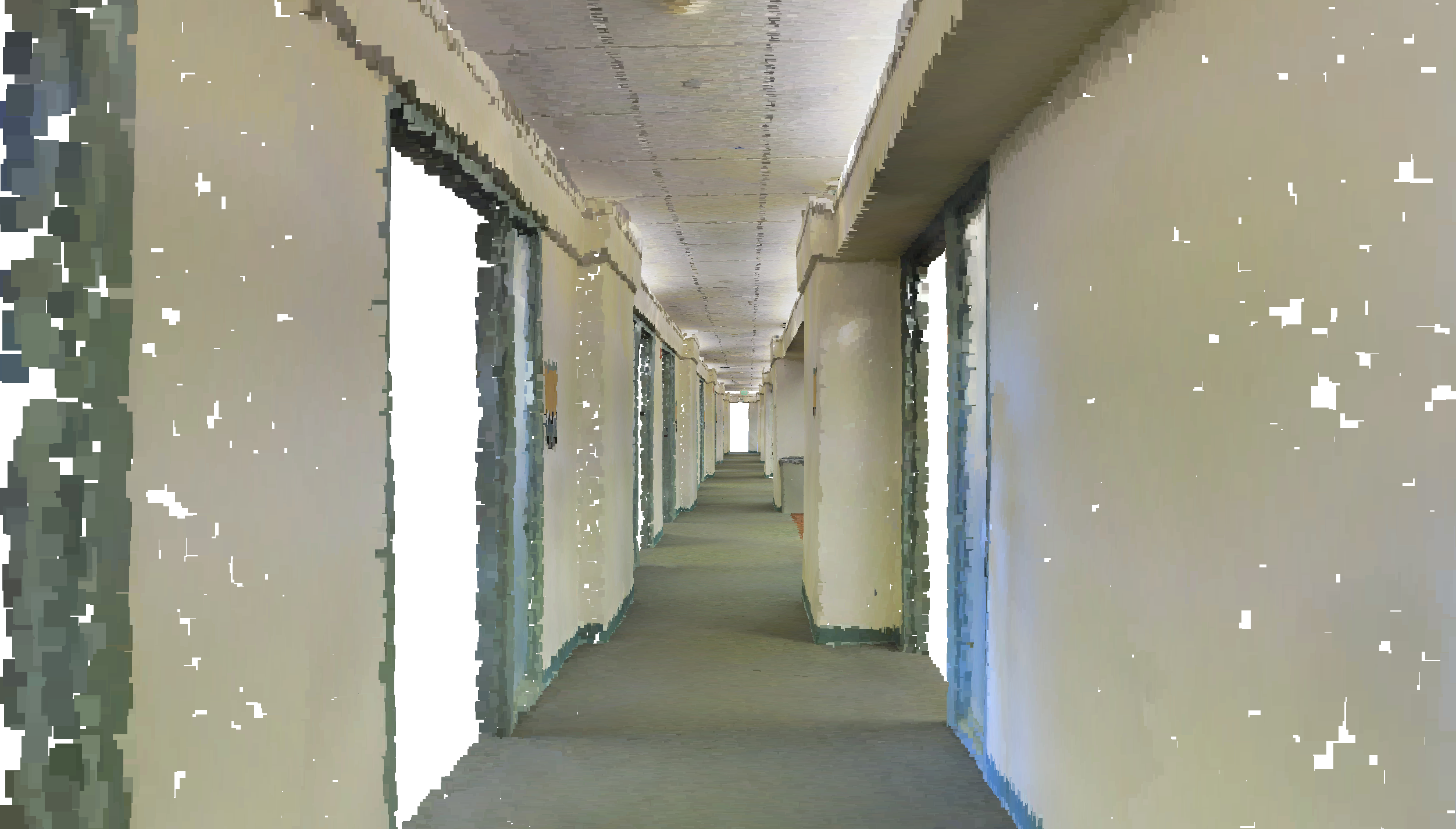}
    \end{subfigure}
    \begin{subfigure}[b]{0.3\textwidth}
        \centering
        \includegraphics[width=\textwidth]{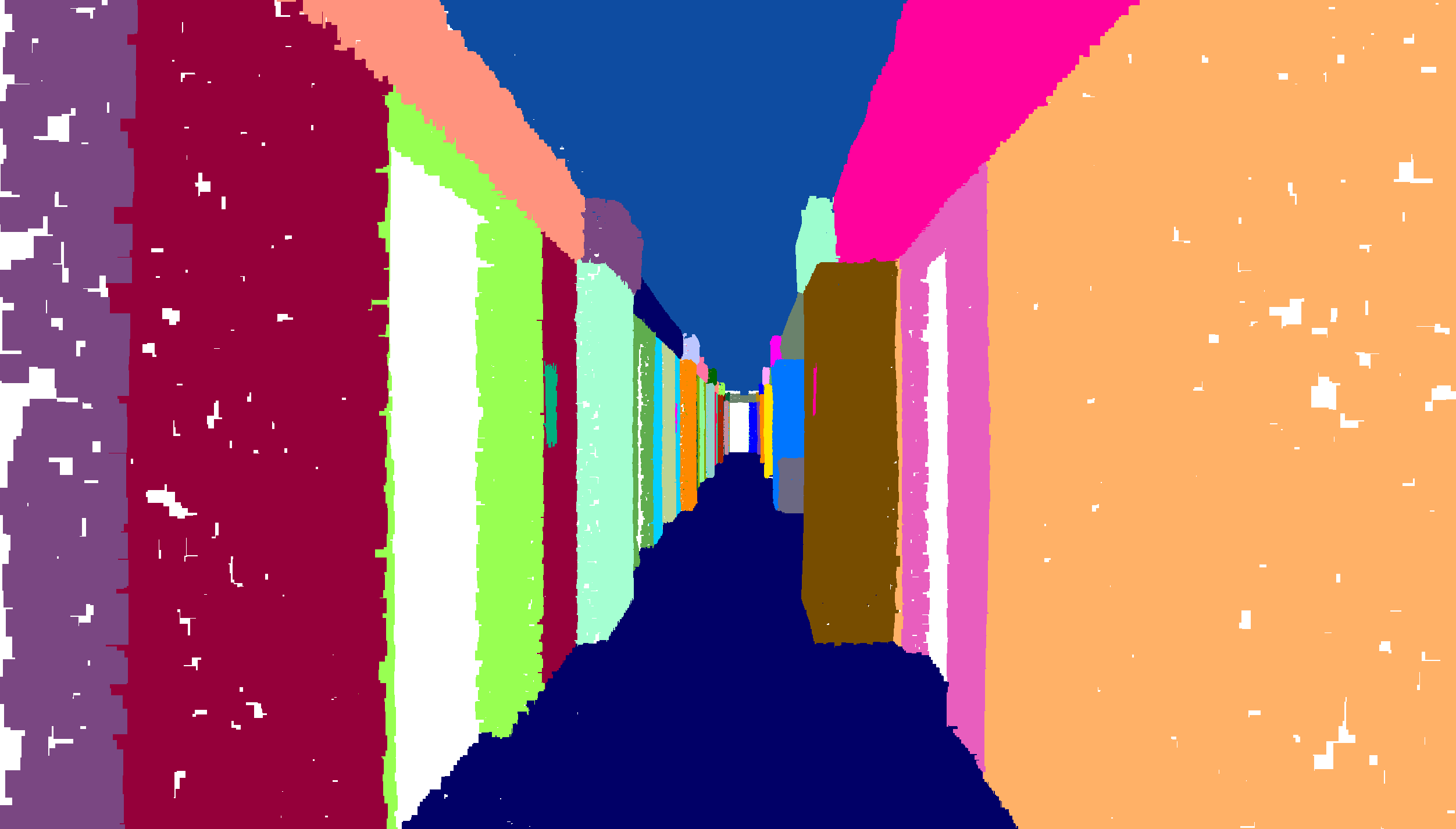}
    \end{subfigure}
    \begin{subfigure}[b]{0.3\textwidth}
        \centering
        \includegraphics[width=\textwidth]{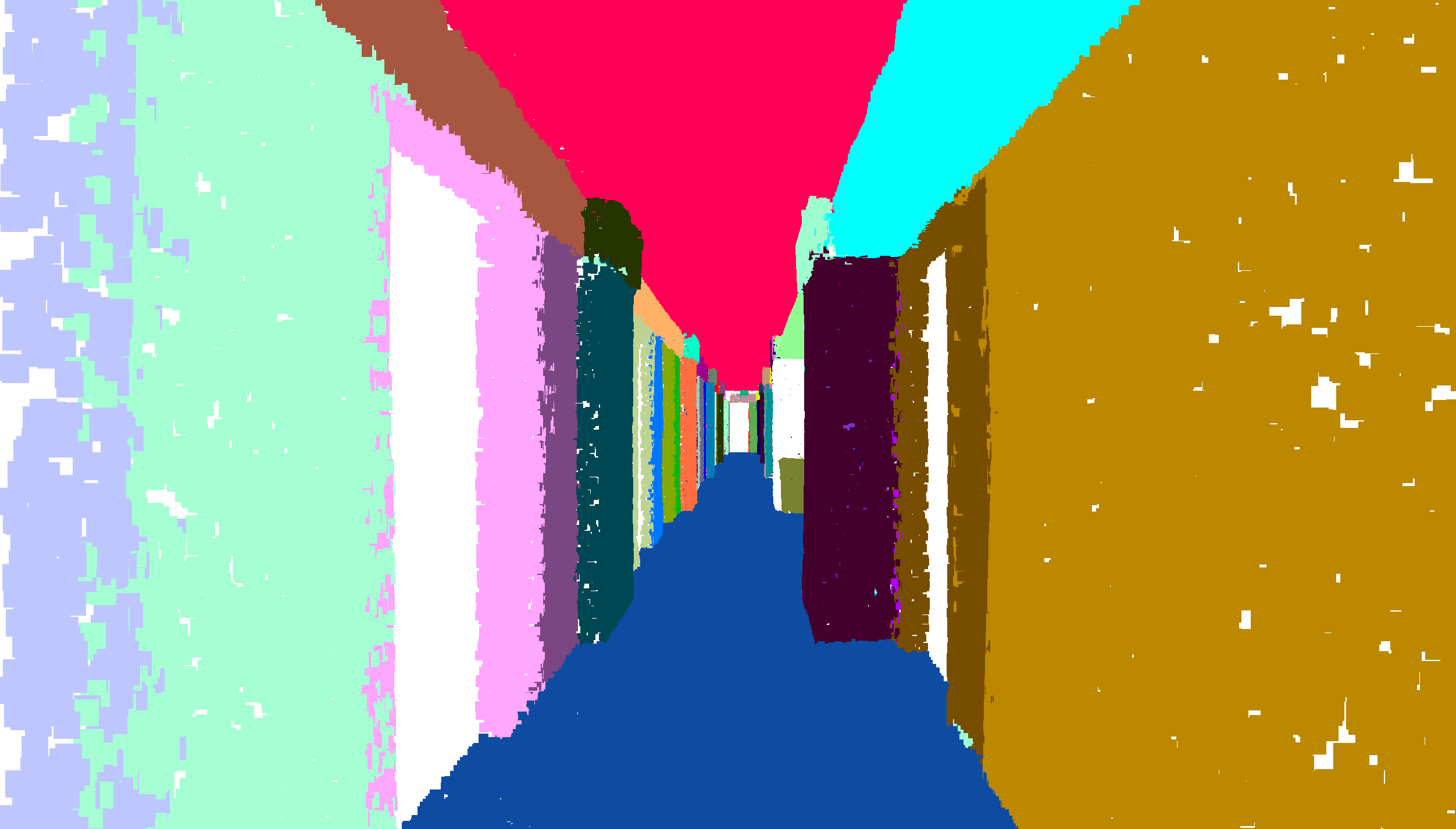}
    \end{subfigure}

    \centering
    \begin{subfigure}[b]{0.3\textwidth}
        \centering
        \includegraphics[width=\textwidth]{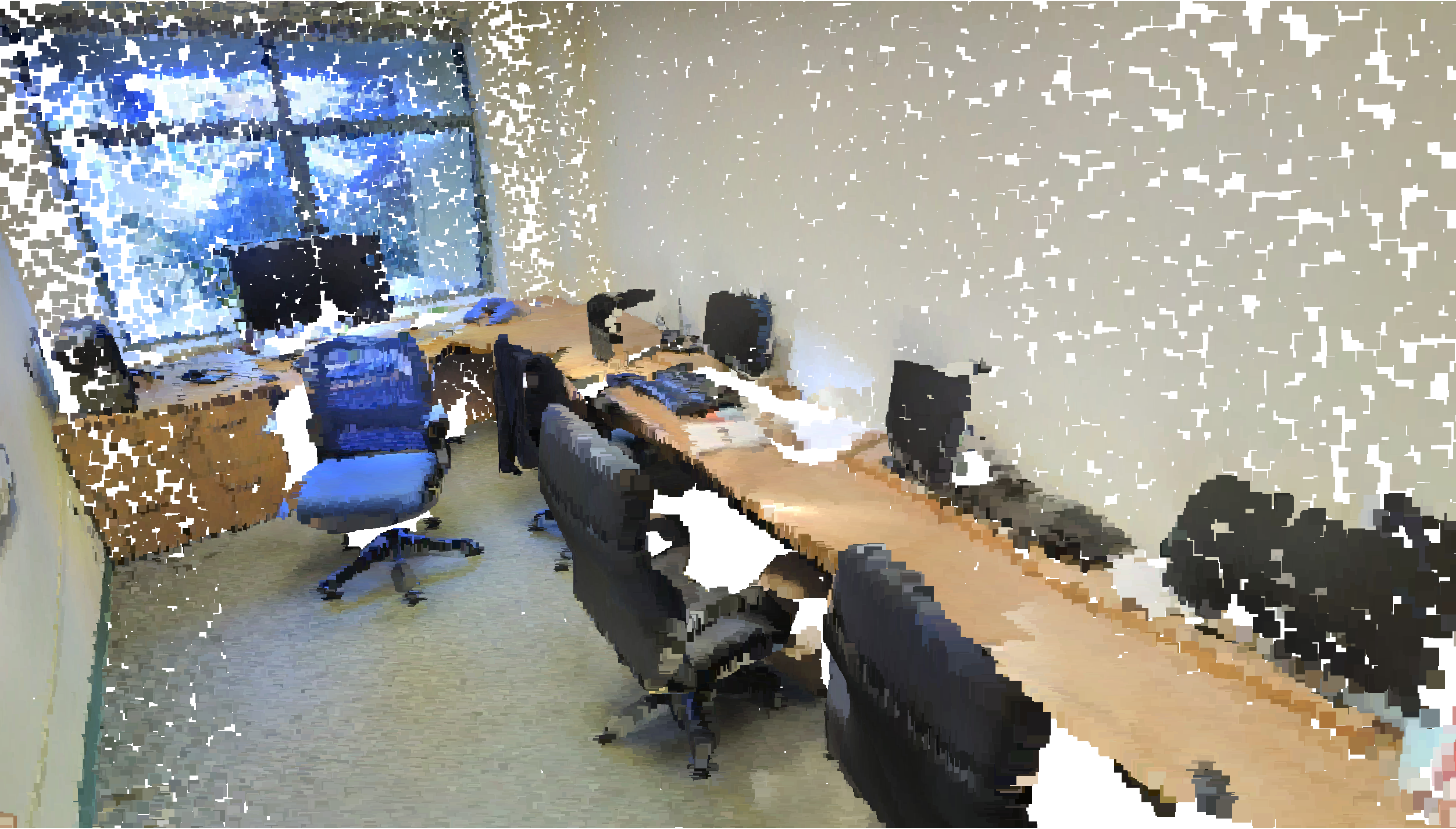}
        \caption{Input}
        \label{subfig:s3dis_vis_input}
    \end{subfigure}
    \begin{subfigure}[b]{0.3\textwidth}
        \centering
        \includegraphics[width=\textwidth]{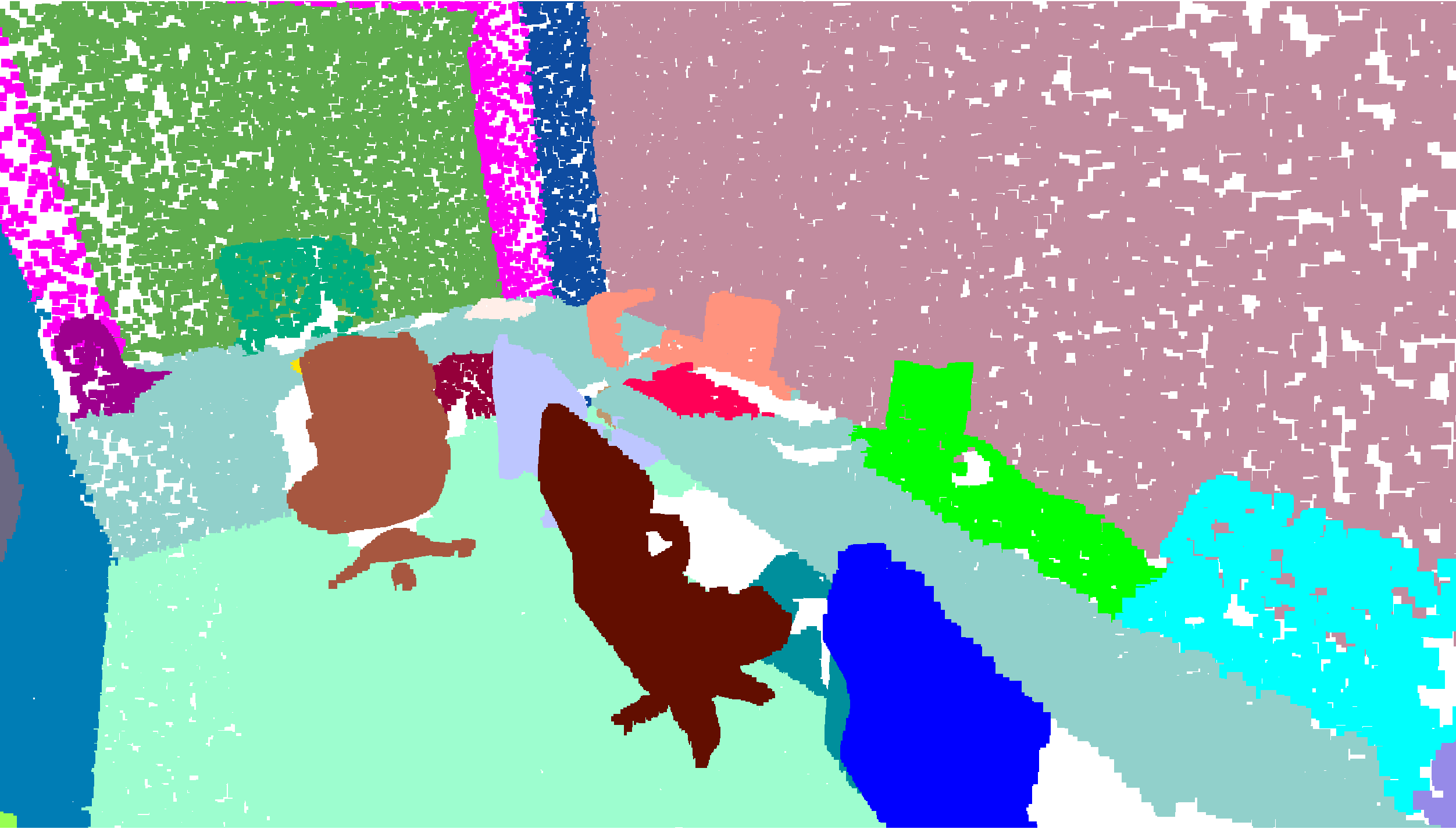}
        \caption{Instance GT}
        \label{subfig:s3dis_vis_gt}
    \end{subfigure}
    \begin{subfigure}[b]{0.3\textwidth}
        \centering
        \includegraphics[width=\textwidth]{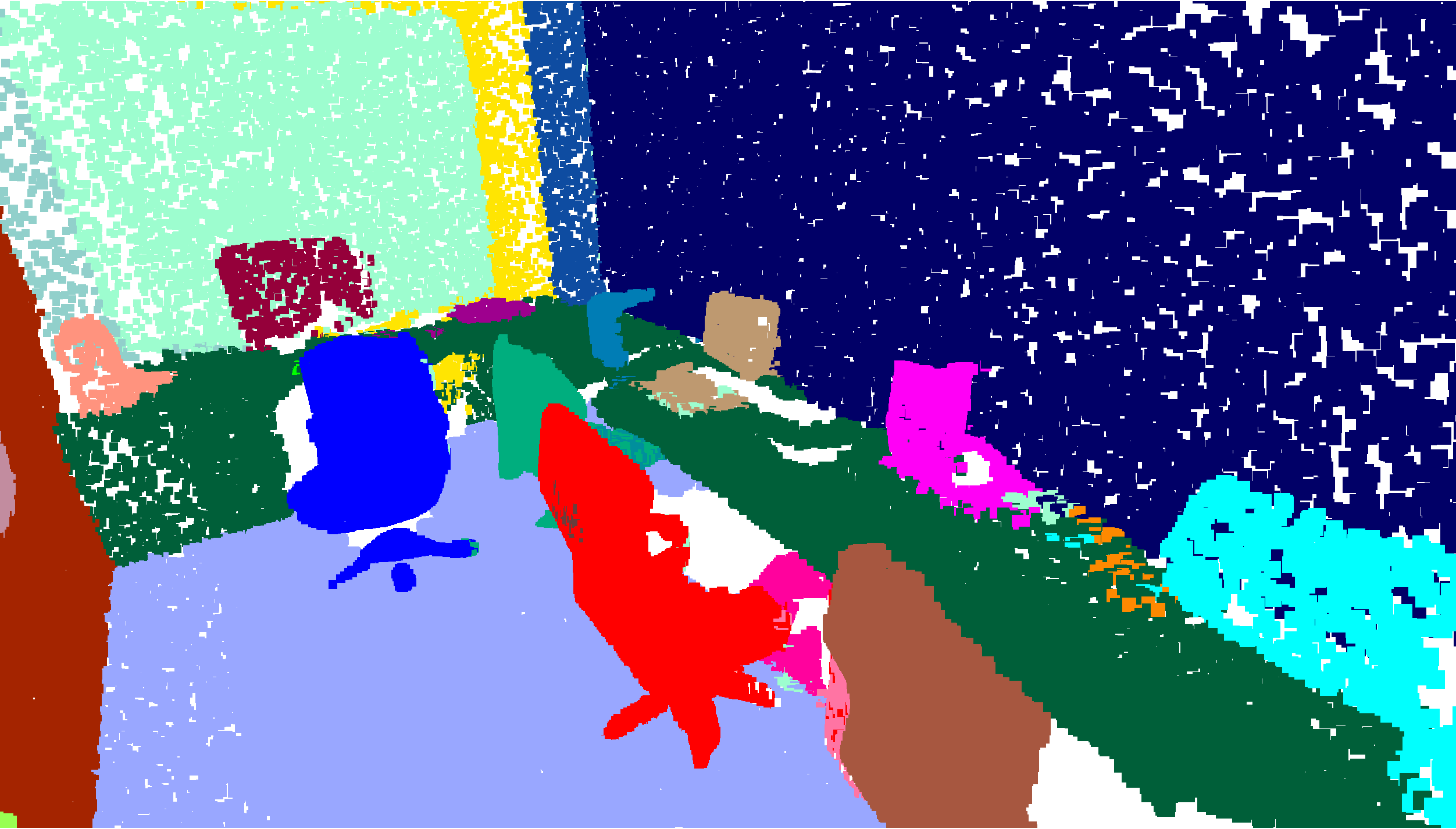}
        \caption{Instance predictions}
        \label{subfig:s3dis_vis_pred}
    \end{subfigure}
    \caption{Visualization of S3DIS results. Note that different colors represent different instances and that the same instance may have a different color in the ground-truth and prediction}
    \label{fig:s3dis_vis}
\end{figure*}

The quantitative results of our method and comparison with the state-of-the-art are shown in \cref{tab:s3dis}. To allow a fair comparison, all full-scene methods have been retrained to operate on blocks using the official, publicly available code. 
It should be noticed that their performance dropped in comparison to their original full-scene setup due to the loss of a global overview. This is especially true for SoftGroup \citep{vu2022softgroup}, which relies on the average number of points per class for candidate selection. Consequently, the presence of partial objects leads to unsatisfactory results when using this particular architecture. 


As shown in \cref{tab:s3dis}, ProtoSeg greatly outperforms all state-of-the-art methods in 7 out of the 8 comparisons. For 5th-fold validation the mean coverage, mean weighted-coverage, mean recall and mean precision are improved with 0.3\%, 3.6\%, 4.9\% and 3.6\%, respectively, when compared to the state-of-the-art. With the exception of the mean coverage, similar results are obtained for 6th-fold cross validation, for which the improvements are 1.1\%, 2.6\%, and 2.4\%. From the experiments, one can safely conclude that the proposed method yields the best overall performance on S3DIS-blocks. 


\begin{table}
\begin{center}
\begin{tabular}{|c|c|c|c||c|} 
    \hline
    Method & Level-1 & Level-2 & Level-3 & Average \\ 
    \hline
    SGPN~\citep{wang2018sgpn} & 72.4 & 25.4 & 19.4 & 39.1\\ 
    PartNet~\citep{mo2019partnet} & 74.4 & 35.5 & 29.0 & 46.3\\ 
    GSPN~\citep{yi2019gspn} & - & - & 26.8 & -\\
    MPNet~\citep{he2020learning} & 79.9 & 41.2 & 32.5 & 51.2\\
    PE~\citep{zhang2021point} & 77.1  & 38.6 & 34.7 & 50.1\\
    IAM~\citep{he2020instance} & 79.5 & 38.6 & 31.2 & 49.7\\
    \hline
    Ours & \textbf{80.9} & \textbf{42.5} & \textbf{36.3} & \textbf{53.2}\\ 
    \hline
\end{tabular}
\caption{Comparison on the Chair category of PartNet. Results are expressed in mAP (IoU threshold of 0.5). The last column is the average mAP of the 3 granularity levels.}
\label{tab:partnet}
\end{center}
\end{table}

Qualitative results are presented in \cref{fig:s3dis_vis}. Overall, one can notice very accurate segmentation. For the corridor nearly all instances are detected. Only the smaller ones, such as the sign next to the openings are missing. Also for the office, segmentation results are very adequate. Chairs, walls and apparel are all detected.

Lastly, the instance segmentation inference time of our method, compared to the state-of-the-art methods, is presented in \cref{tab:inference_time}. Important to notice is that our method is not only the fastest overall, a reduction in inference time of 28\% is realized, but also achieves the lowest standard deviation. While the state-of-the-art achieves a standard deviation in the range of 10.8-53.1\%, ours is only 1.0\% of the total time. This time predictability is important for real-time applications and results from our overcomplete approach which is independent of the number of instances in the underlying scene and the omission of the clustering step.

\begin{figure*}
    \centering
    \begin{subfigure}[b]{0.150\textwidth}
        \centering
        \includegraphics[width=\textwidth]{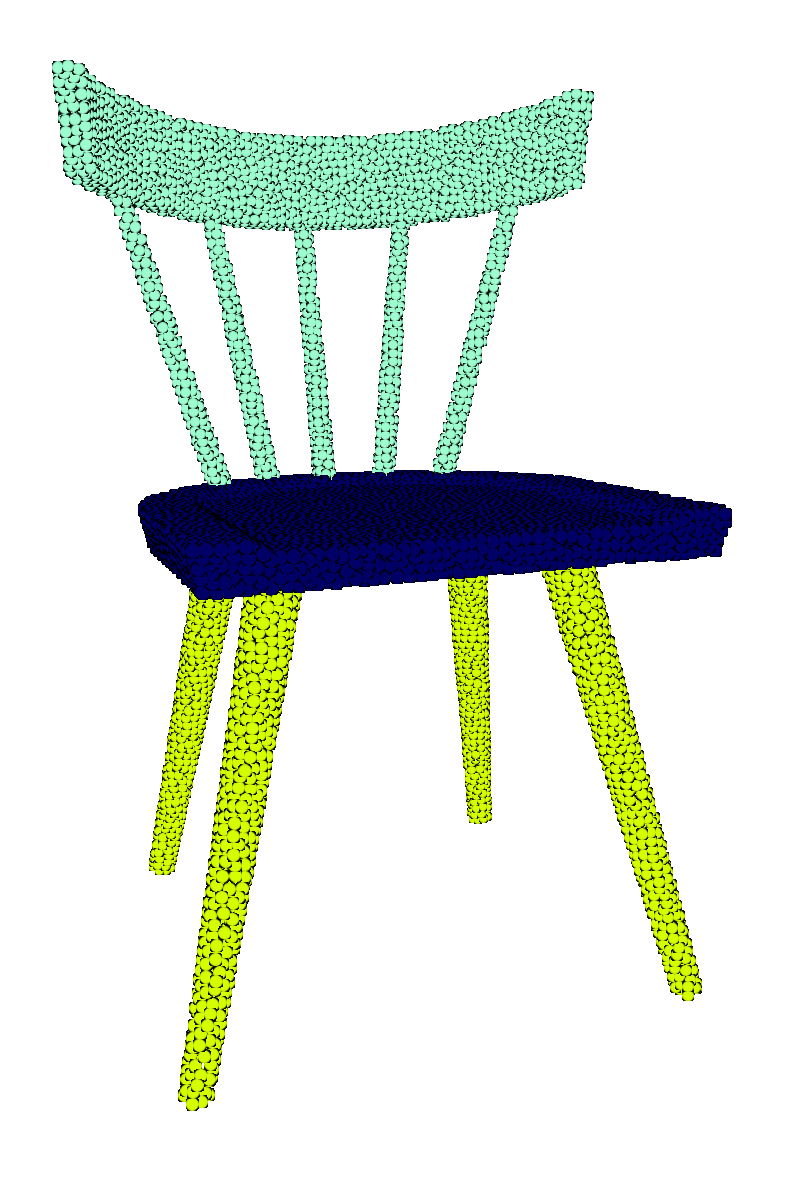}
    \end{subfigure}
    \begin{subfigure}[b]{0.150\textwidth}
        \centering
        \includegraphics[width=\textwidth]{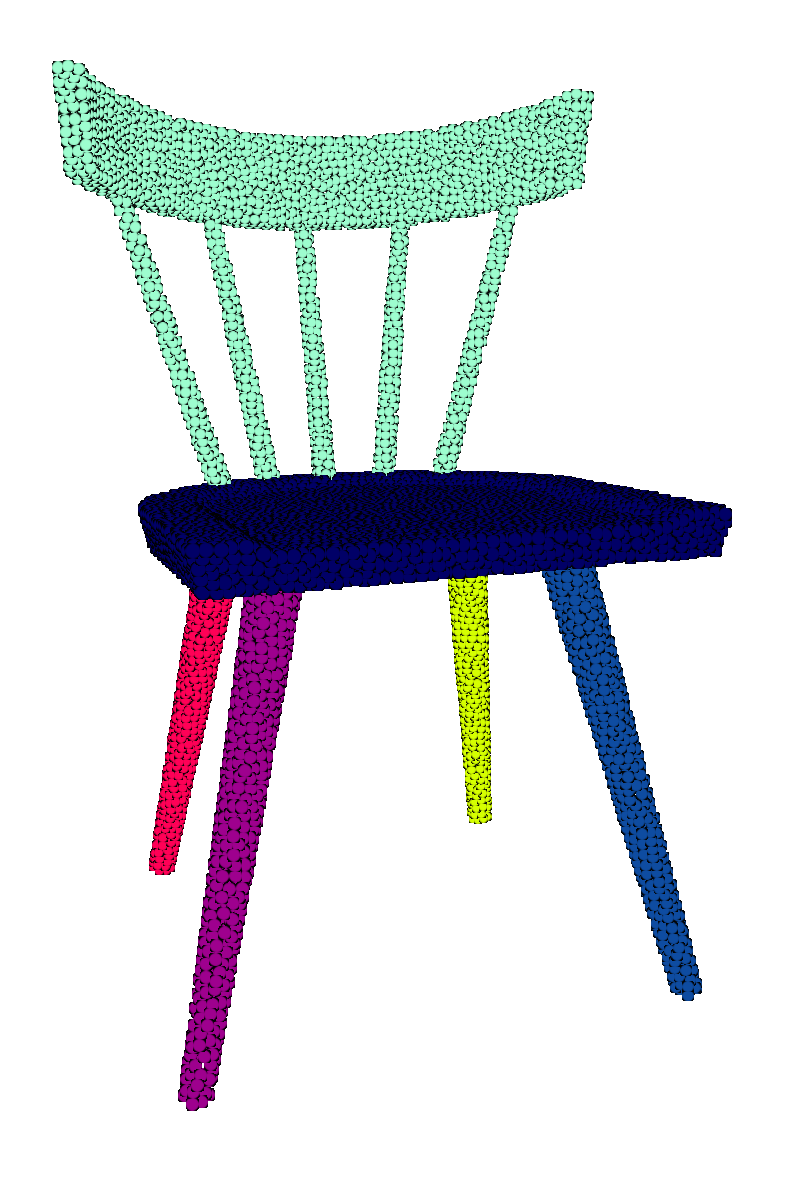}
    \end{subfigure}
    \begin{subfigure}[b]{0.150\textwidth}
        \centering
        \includegraphics[width=\textwidth]{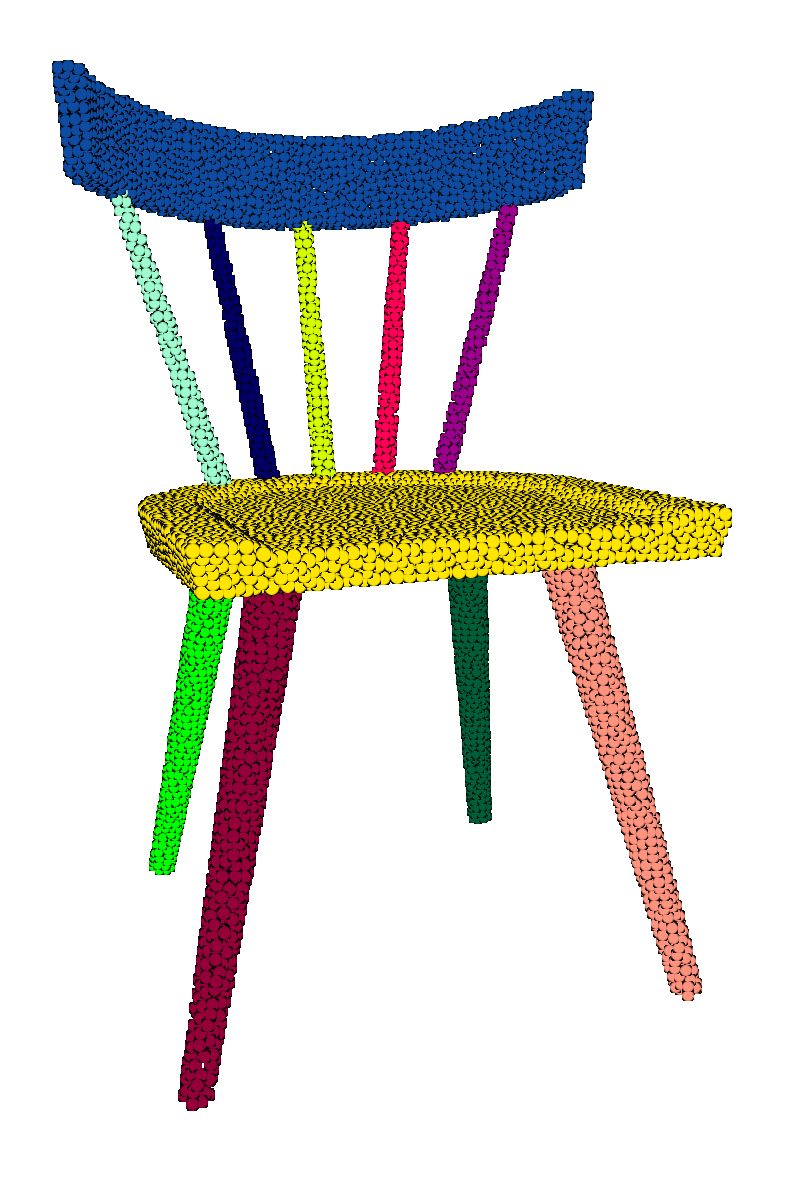}
    \end{subfigure}
    \begin{subfigure}[b]{0.150\textwidth}
        \centering
        \includegraphics[width=\textwidth]{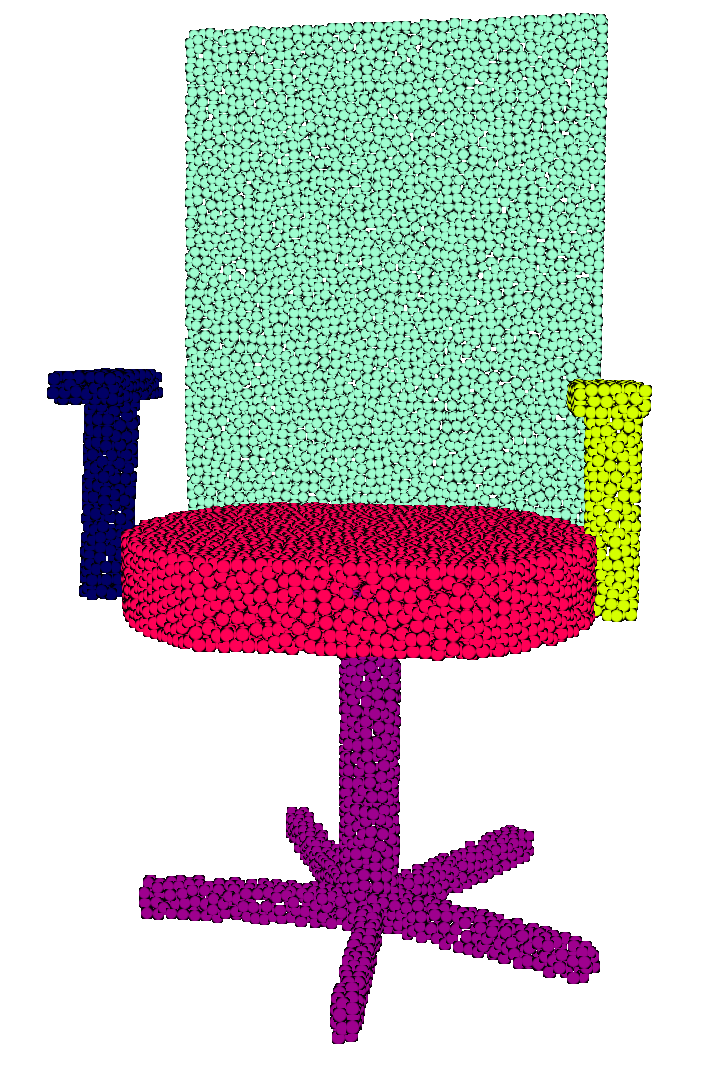}
    \end{subfigure}
    \begin{subfigure}[b]{0.150\textwidth}
        \centering
        \includegraphics[width=\textwidth]{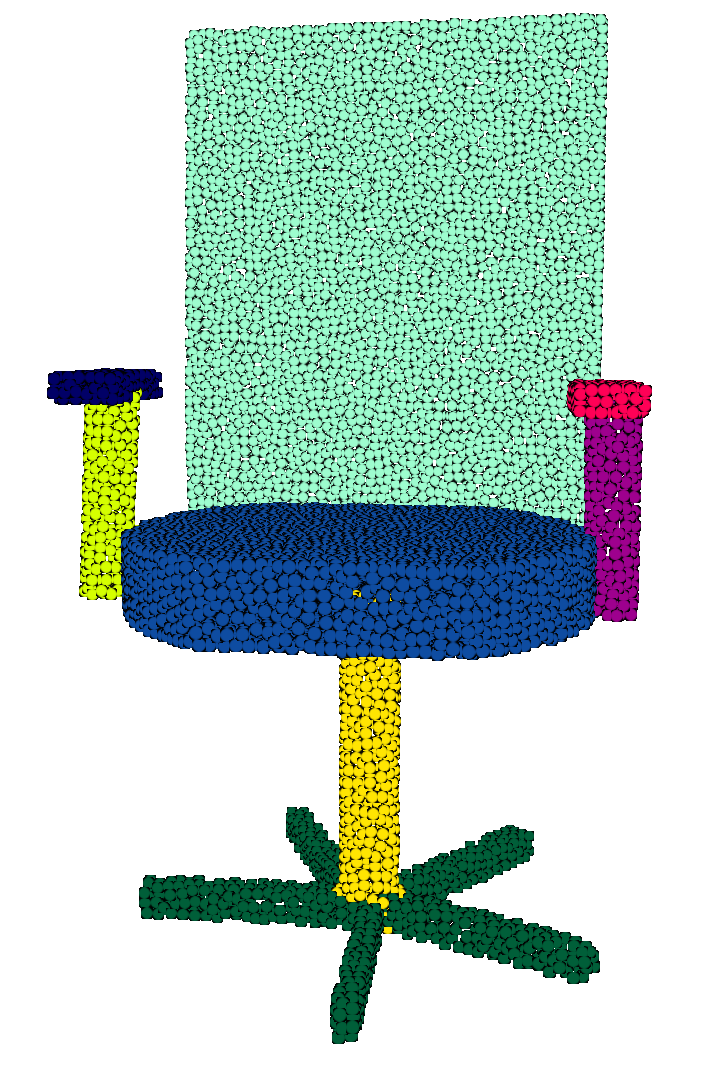}
    \end{subfigure}
    \begin{subfigure}[b]{0.150\textwidth}
        \centering
        \includegraphics[width=\textwidth]{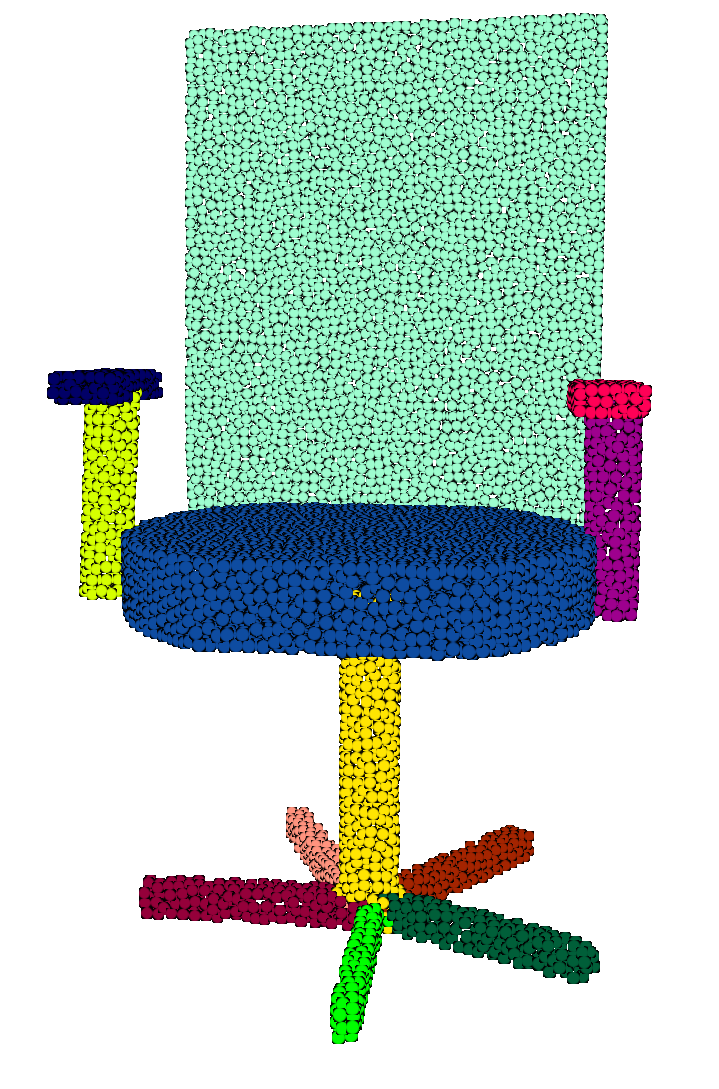}
    \end{subfigure}

    \centering
    \centering
    \begin{subfigure}[b]{0.150\textwidth}
        \centering
        \includegraphics[width=\textwidth]{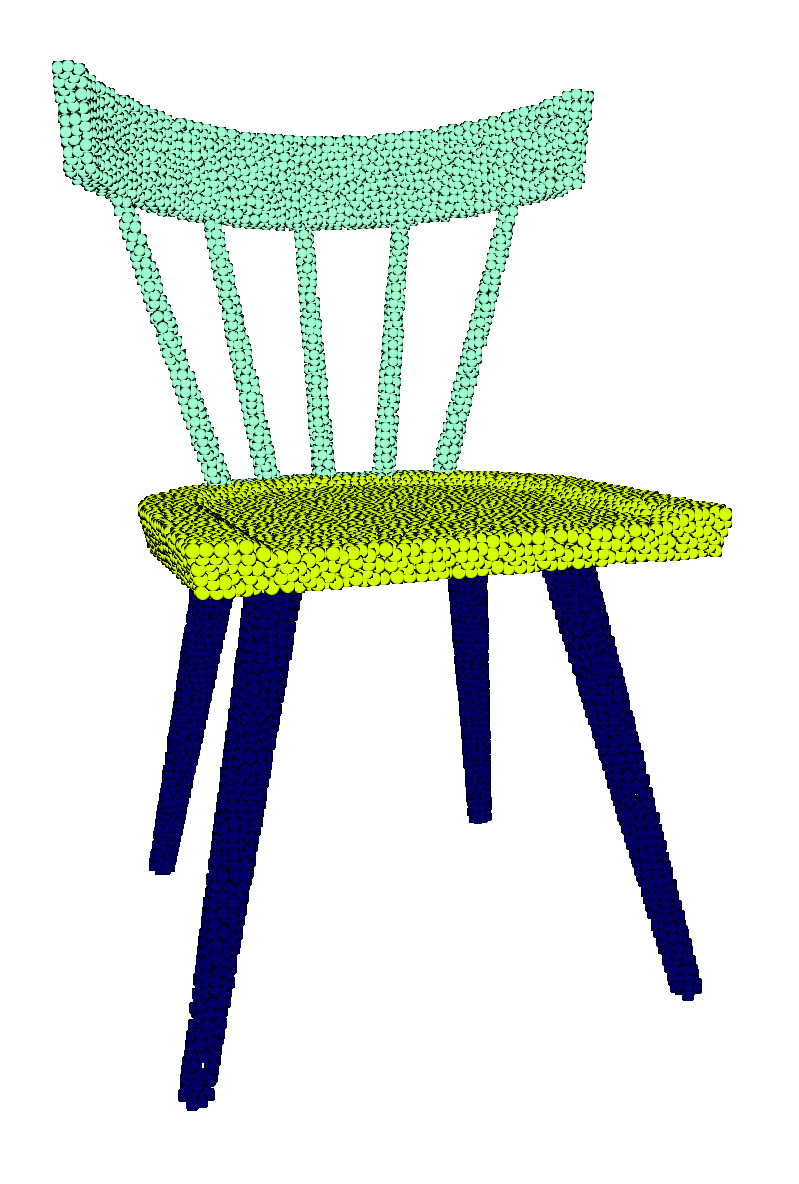}
    \end{subfigure}
    \begin{subfigure}[b]{0.150\textwidth}
        \centering
        \includegraphics[width=\textwidth]{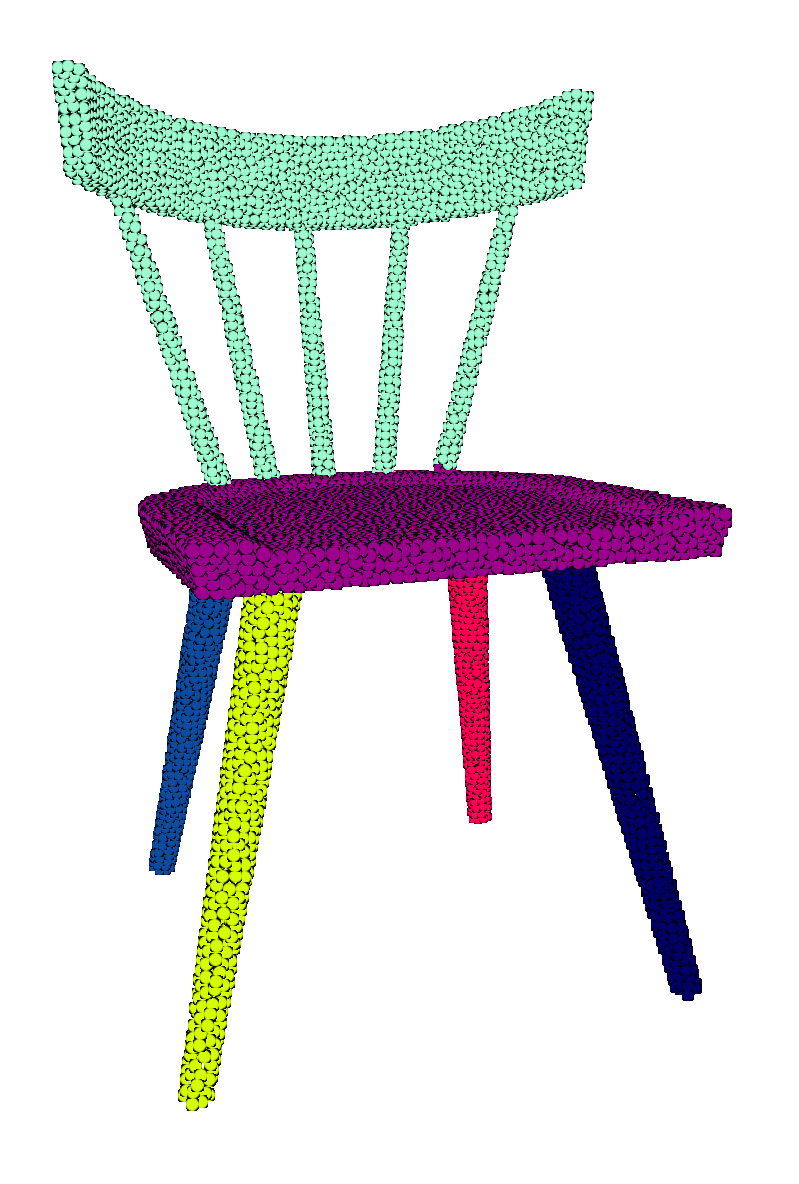}
    \end{subfigure}
    \begin{subfigure}[b]{0.150\textwidth}
        \centering
        \includegraphics[width=\textwidth]{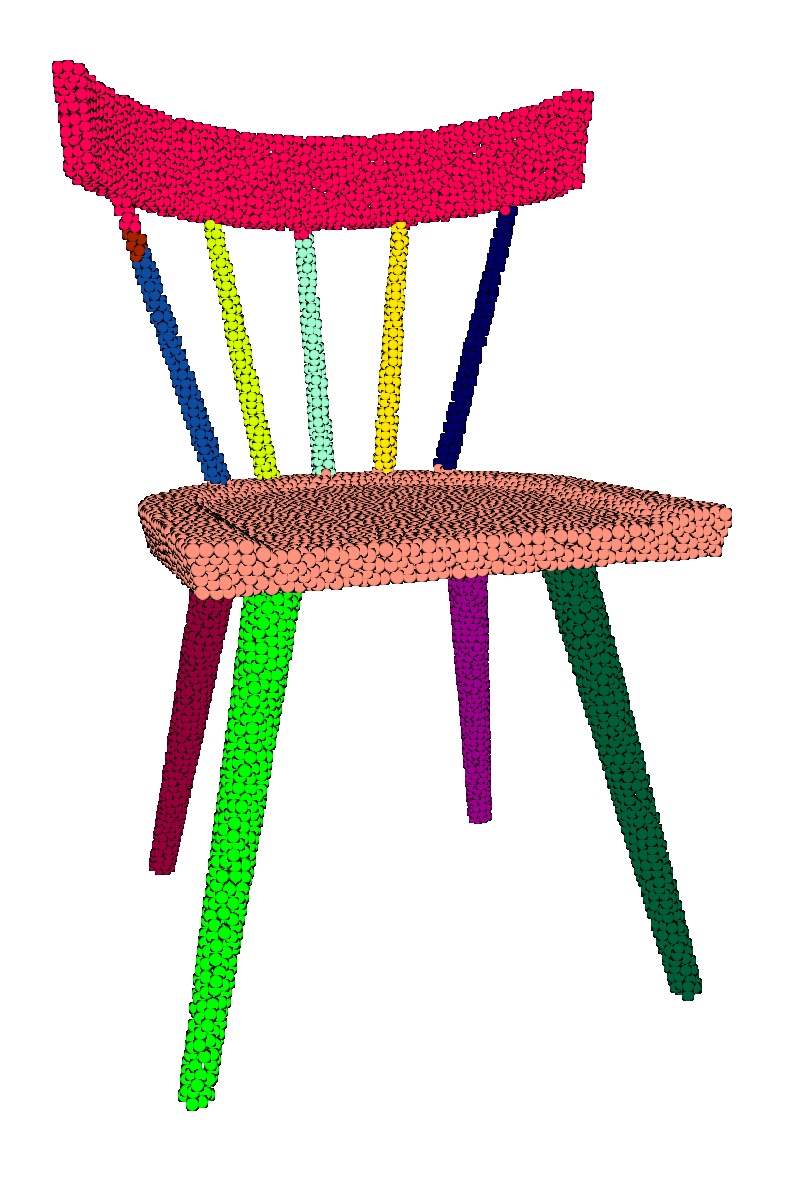}
    \end{subfigure}
    \begin{subfigure}[b]{0.150\textwidth}
        \centering
        \includegraphics[width=\textwidth]{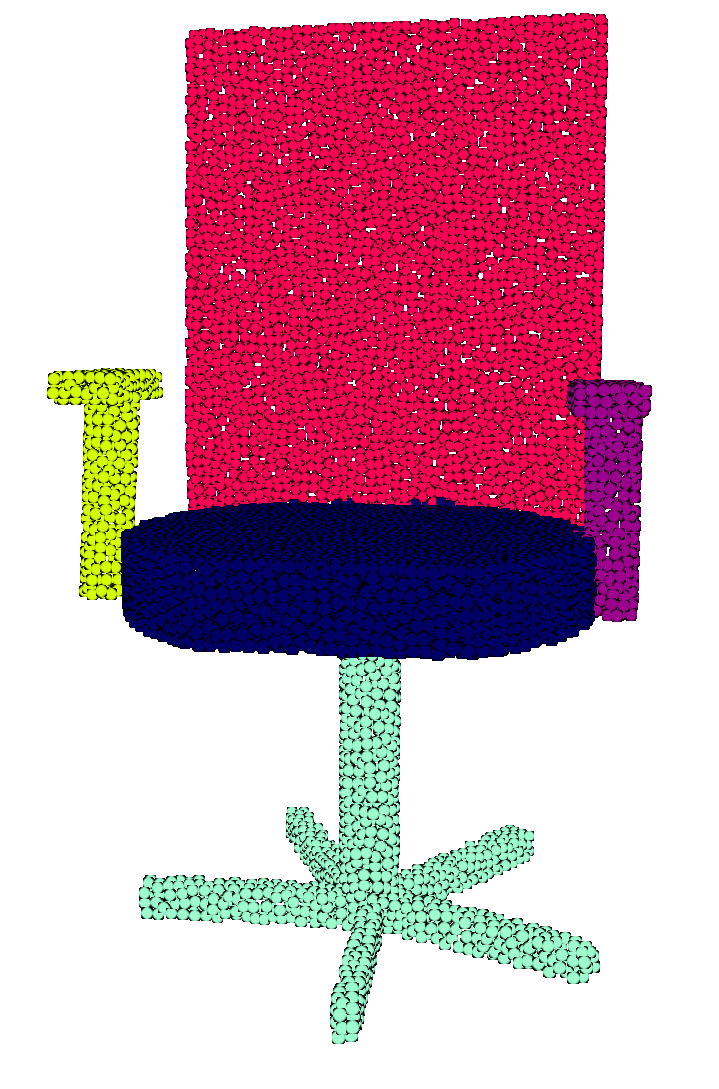}
    \end{subfigure}
    \begin{subfigure}[b]{0.150\textwidth}
        \centering
        \includegraphics[width=\textwidth]{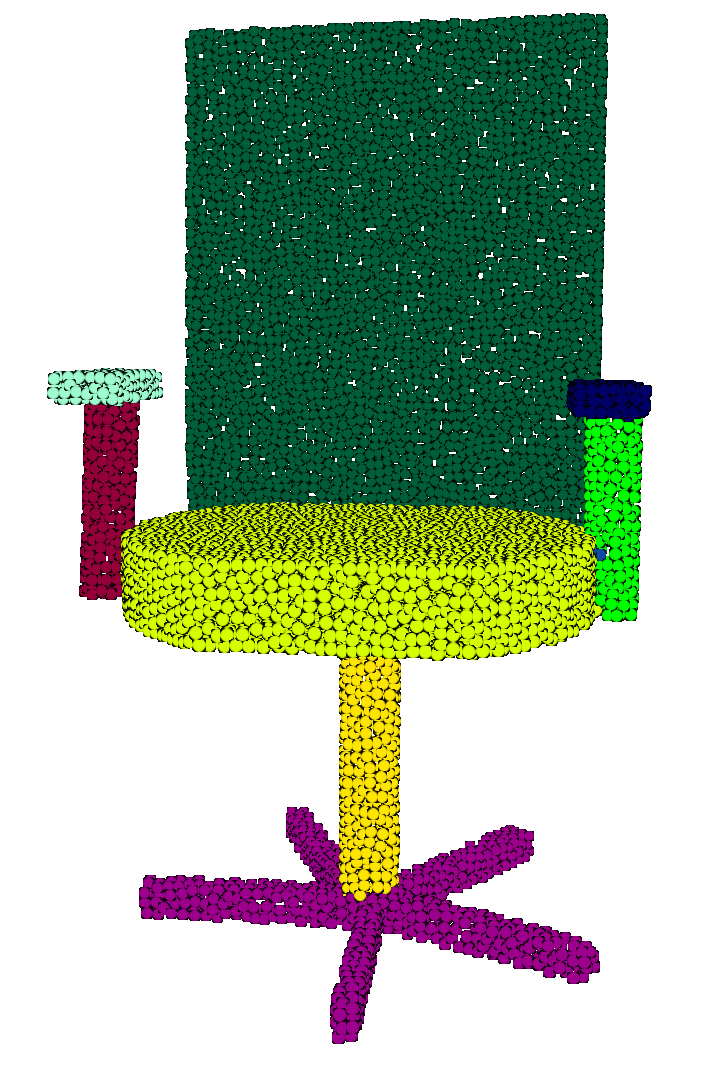}
    \end{subfigure}
    \begin{subfigure}[b]{0.150\textwidth}
        \centering
        \includegraphics[width=\textwidth]{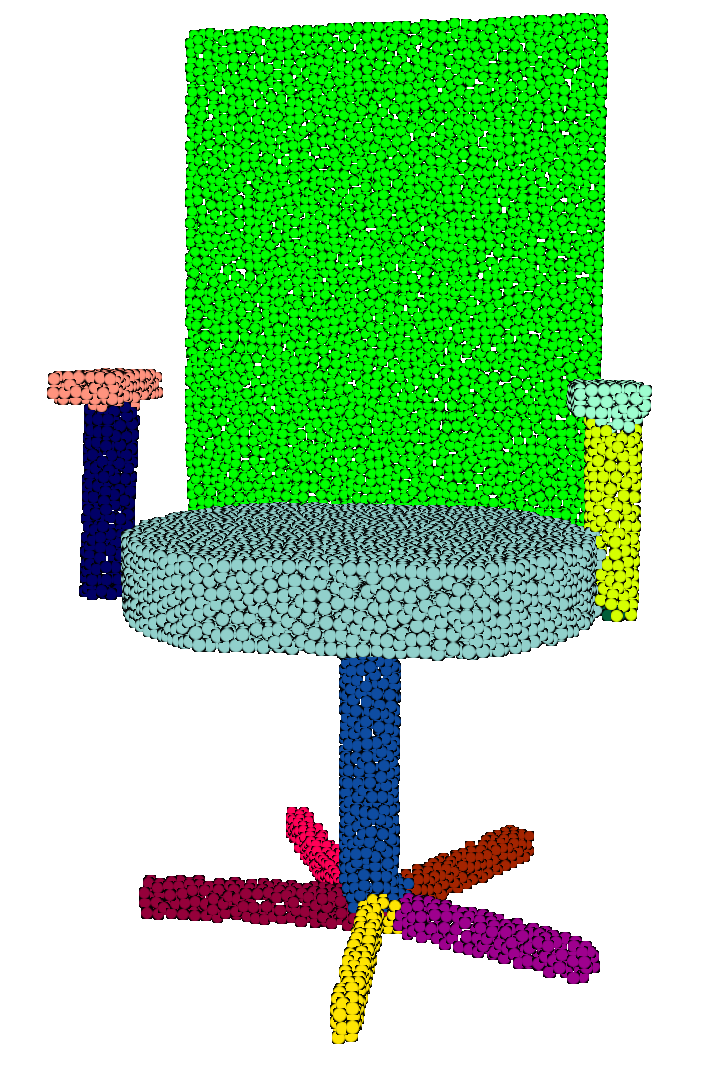}
    \end{subfigure}
    \caption{Two examples point clouds from PartNet at different instance segmentation granularities. The ground truth and predictions are shown on the top and bottom rows, respectively.}
    \label{fig:partnet_vis}
\end{figure*}

\subsection{PartNet Part Instance Segmentation}
To show the generality of the proposed method, we also apply it on a dataset which fundamentally differs from S3DIS. While PartNet consists of objects with more complex shapes, we are still able to achieve state-of-the-art results. Quantitative results are provided in \cref{tab:partnet}. For this experiment, ProtoSeg offers state-of-the-art performance for all three granularity levels. Moreover, on average, we outperform the second best with 2.0\%. Qualitative results are shown in \cref{fig:partnet_vis}. Note that for all granularity levels a consistent and refined segmentation is achieved.


\subsection{Ablation study}
\label{subsec:ablation}

To validate the design of ProtoSeg, we perform an ablation study using the Area-5 S3DIS data and evaluate on mRec and mPrec. Specifically, we study the effect of our novel DPI module and reciprocal loss function. The results can be found in \cref{tab:ablation study}. The employment of the DPI module instead of a single-scale PointConv layer leads to an improvement of 1.6\% in mPrec. The addition of the second term in the reciprocal loss has an even larger effect as it allows an improvement of 2.2\% in mPrec and 1.8\% in mRec. Thus, the ablation clearly proves the importance of these two additions.

%% file: texts/5_conclusion.tex
\section{Conclusions}

This paper presents a novel deep neural network for 3D instance segmentation. The proposed architecture generates prototypes and jointly learns their associated coefficients. By linearly combining those, the resulting instance masks are obtained. Next, non-maximum suppression is employed to yield the final prediction. In order to retrieve multi-scale coefficients using an overcomplete set of deterministically sampled points, we introduced a dilated point inception module. A specific reciprocal loss has been designed to enhance performance. Our experiments do not only reveal that the proposed method significantly and consistently outperforms the state-of-the-art for S3DIS-blocks~\citep{armeni20163d} (4.9\% in mRec on Fold-5) and PartNet~\citep{mo2019partnet} (2.0\% on average in mAP) but our clustering-free approach also yields the fastest inference with a reduction of 28\%, compared to the currently fastest method. Our overcomplete approach allows an extremely low standard deviation on the inference time with only 1.0 \% of the total time, in comparison to the 10.8- 53.1\% range of the state-of-the-art.

\begin{table}
\begin{center}
\begin{tabular}{|c|c|c||c|c|} 
    \hline
    \multirow{2}{*}{DPI} & \multicolumn{2}{c||}{Reciprocal loss} & \multirow{2}{*}{mRec} & \multirow{2}{*}{mPrec} \\
    \cline{2-3}
     & $J_{PR \rightarrow GT}$ & $J_{GT \rightarrow PR}$ & & \\
    \hline
    \cmark &\cmark & \cmark & \textbf{54.5} & \textbf{66.1}\\ 
    \xmark &\cmark & \cmark & 54.4 & 64.5\\
    \cmark &\cmark & \xmark & 52,7 & 63,9\\ 
    \hline
\end{tabular}
\caption{Ablation study of DPI module and reciprocal loss terms.}
\label{tab:ablation study}
\end{center}
\end{table}